\documentclass[runningheads]{llncs}

\usepackage[mobile]{eccv}

\usepackage{graphicx}
\usepackage{amsmath}
\usepackage{amssymb}
\usepackage{booktabs}
\usepackage{xcolor}
\usepackage{float}
\usepackage{caption}
\usepackage{multirow}
\usepackage{eccvabbrv}
\usepackage{microtype}
\usepackage{graphicx}
\usepackage{booktabs}
\usepackage{colortbl}
\usepackage[accsupp]{axessibility}  

\usepackage[pagebackref,breaklinks,colorlinks,citecolor=eccvblue]{hyperref}

\usepackage{orcidlink}

\begin{document}

\title{DeepSport: A Multimodal Large Language Model for Comprehensive Sports Video Reasoning via Agentic Reinforcement Learning}
\titlerunning{DeepSport}

\author{Junbo Zou\inst{1}$^{*}$ \and
Haotian Xia\inst{2}$^{*}$ \and
Zhen Ye\inst{3} \and
Shengjie Zhang\inst{4} \and
Christopher Lai\inst{5} \and
Vicente Ordonez\inst{2} \and
Weining Shen\inst{4} \and
Hanjie Chen\inst{2}
}

\authorrunning{J. Zou, H. Xia, et al.}

\institute{$^{1}$Georgia Institute of Technology \quad
$^{2}$Rice University \quad
$^{3}$Johns Hopkins University\\
$^{4}$University of California, Irvine \quad
$^{5}$University of California, Santa Barbara\\
jzou98@gatech.edu, \{haotian.xia, hanjie\}@rice.edu
}

\maketitle
\renewcommand\thefootnote{}
\footnotetext[1]{* Equal contribution.}

\begin{abstract}
Sports video understanding requires perceiving high-speed dynamics, complex rules, and long temporal contexts. Yet, current Multimodal Large Language Models (MLLMs) remain narrowly focused on single sports, specific tasks, or training-free paradigms. We introduce \textbf{DeepSport}, the first end-to-end trained MLLM for multi-task, multi-sport video understanding. DeepSport shifts from passive frame processing to active, iterative reasoning, dynamically extracting frames to ``think with videos.'' To train our model, we curate a unified 78k-sample dataset via a rigorous three-step text-and-vision distillation pipeline. We then employ a progressive two-stage training strategy: a Sports Curriculum Supervised Fine-Tuning phase to build foundational perception, followed by Agentic Reinforcement Learning with a novel tool-use reward. Extensive experiments on a comprehensive 6.7k benchmark demonstrate that DeepSport achieves state-of-the-art performance—outperforming powerful proprietary and open-source models—while utilizing significantly fewer frames. Furthermore, it exhibits strong zero-shot transferability to unseen sports and broad motion recognition tasks, establishing a highly efficient and generalized foundation for complex video reasoning.
\end{abstract}    

\section{Introduction}
\label{sec:intro}

\begin{figure*}
    \centering
    \includegraphics[width=0.85\linewidth]{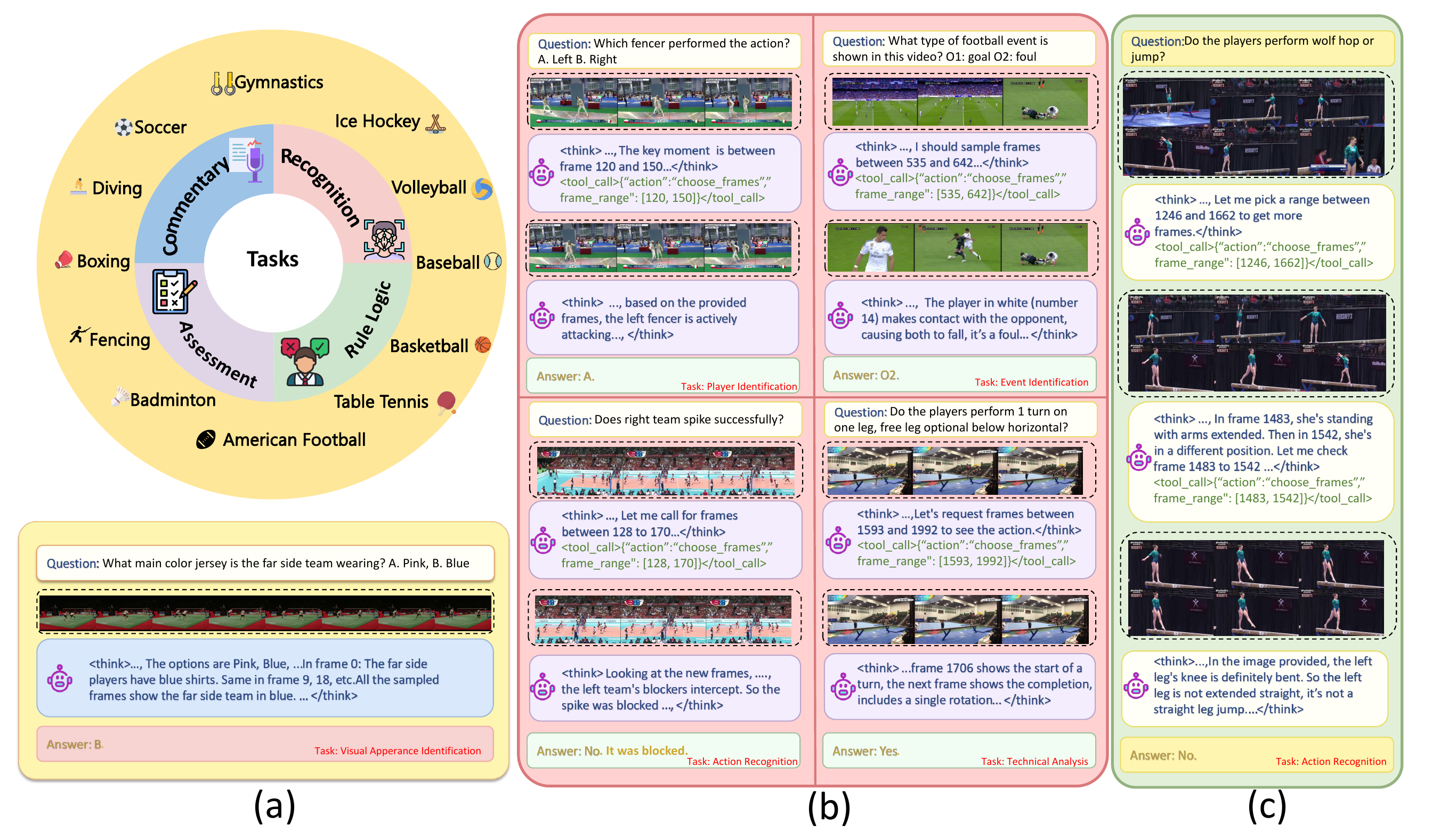}
    \caption{Overview of the DeepSport framework. Trained through SFT and RL, DeepSport decides whether to utilize tools, supporting (a) passive single-pass inference, (b) single tool-use for two-turn interactions, and (c) iterative multi-turn reasoning via multiple tool calls. As shown on the Left, we categorize tasks into four core dimensions: Fine-Grained Recognition, Rule \& Procedural Logic, Assessment \& Coaching, and Live Commentary, covering diverse fine-grained sub-tasks (in red) across multiple sports.}
    \label{fig:teaser}
\end{figure*}

The domain of sports analysis has long served as a fertile ground for interdisciplinary research, driving advancements in both computer vision and natural language processing~\cite{THOMAS20173}. Traditional vision approaches have achieved remarkable success in specific, isolated tasks, such as fine-grained action detection and recognition~\cite{shao2020finegym, giancola2018soccernet}, event spotting~\cite{Xarles_2024_CVPR}, highlight generation~\cite{merler2018excitement, merler2018automatic}, and player and ball trajectory detection and tracking in team sports~\cite{cioppa2021cameracalibrationplayerlocalization, vandeghen2022semi, 10411638,cite-key}. Simultaneously, NLP techniques have been effectively deployed for automated match reporting~\cite{huang2020generating} and sentiment analysis~\cite{vujivcic2023approach}. Recently, the emergence of Multimodal Large Language Models (MLLMs)~\cite{openai2024gpt4ocard, geminiteam2024gemini} has accelerated this trend, enabling more complex applications in sports~\cite{xia2024language}, such as AI-assisted refereeing~\cite{held2023vars}, trajectory generation~\cite{capellera2026jointdiffbridgingcontinuousdiscrete}, and commentary generation~\cite{rao2024matchtimeautomaticsoccergame, you2025timesoccer}. 

In response to these capabilities, the community has produced a surge of high-quality, public benchmarks covering a wide array of disciplines. These range from action quality assessment in diving~\cite{xu2022finediving} and fine-grained player skill estimation in basketball~\cite{pan2025basket},  to tactical classification in fencing~\cite{lai2024facts}, and comprehensive question answering in large-scale benchmarks~\cite{xia2024sportqa,xia2024sportu,rao2025multi}. These datasets provide the necessary foundation for evaluating sophisticated sports intelligence.

Despite the creation of data and benchmarks, the development of capable models has lagged behind. Current sports-specific MLLMs exhibit a significant fragmentation. For example, trained models are predominantly soccer-centric~\cite{rao2025multi, you2025timesoccer}, optimizing for specific tasks such as commentary generation within a single sport, but failing to generalize to others. On the other hand, attempts at multi-sport reasoning, such as FineQuest~\cite{chen2025finequestadaptiveknowledgeassistedsports}, rely on training-free paradigms. Consequently, to the best of our knowledge, a unified, end-to-end trained MLLM capable of performing multiple tasks (e.g., commentary, foul detection, QA) across diverse sports does not yet exist.

To bridge this gap, we introduce \textbf{DeepSport}, the first sports-specific MLLM framework explicitly trained for multi-task, multi-sport video understanding (Figure~\ref{fig:teaser}). Inspired by the emerging paradigm of ``Thinking with Videos''~\cite{zhang2025thinkingvideosmultimodaltoolaugmented, he2025framethinkerlearningthinklong, ge2025framemindframeinterleavedvideoreasoning}, DeepSport shifts from passive frame processing to active, iterative video interrogation via a specialized frame-extraction tool. To enable this, we devise a rigorous data distillation pipeline that synthesizes high-quality Chain-of-Thought (CoT) trajectories from 10 diverse data sources covering 12 different sports. We then employ a progressive two-stage strategy—a \textbf{Sports Curriculum Supervised Fine-Tuning (SFT)}, inspired by the nature of sports, to incrementally build foundational perception, followed by \textbf{Agentic Reinforcement Learning} with a novel gated reward function—to teach the model how to reason and when to use tools effectively. Our main contributions are summarized as follows:
\begin{itemize}
    \item \textbf{The First Multi-Task, Multi-Sport Trained MLLM.} We propose DeepSport, a unified model framework that generalizes across diverse sports and four core capabilities: Fine-Grained Recognition, Rule \& Procedural Knowledge, Assessment \& Coaching, and Live Commentary \& Reporting. It represents a paradigm shift from static, single-sport models to dynamic, tool-augmented sports intelligence.
    
    \item \textbf{A New Benchmark and Rigorous Distillation Pipeline.} We introduce a pipeline to curate, unify, and distill data from 10 existing sources across 12 sports. By utilizing a stringent three-step LLM/VLM filtering mechanism to guarantee logic and retrieval consistency, we construct 78K high-quality training trajectories and a testing benchmark with 6.7K samples.
    
    \item \textbf{Curriculum SFT and Tool-Based RL Strategy.} Inspired by the nature and specificity of sports, we design a progressive curriculum-learning strategy during SFT to build visual grounding before high-level logic. This is followed by a specialized GRPO reinforcement learning framework with a gated tool-use reward function that incentivizes active, multi-turn visual reasoning.
    
    \item \textbf{SOTA Performance and Broad Generalizability.} DeepSport achieves an average score of 37.67, establishing a new SOTA for sports video tasks. Furthermore, it demonstrates robust zero-shot transferability to unseen sports and broad motion recognition tasks using significantly fewer frames, proving our paradigm internalizes underlying athletic mechanics rather than merely overfitting to specific rules.
\end{itemize}

\section{Related Work}
\label{sec:related_work}
Analyzing sports is fundamentally a dynamic, temporal reasoning problem. Core tasks, such as foul detection in basketball (e.g., traveling) or technical scoring in diving and gymnastics, are defined by sequences of motion over time, not by isolated static image. Consequently, sports analysis necessitates models that can parse and reason over complex video streams.

In this domain, early sports video analysis relied on task-specific, traditional computer vision (CV) pipelines. For example, for player detection and tracking~\cite{cioppa2021cameracalibrationplayerlocalization,liu2023automatedplayeridentificationindexing, cioppa2022soccernet}, rally prediction \cite{xia2022vren, 10411638}, automated commentary generation\cite{huang2020generating, 10871604,10.1145/3341105.3374063}, fine-grained action recognition in specific sports \cite{shao2020finegym, wu2022survey,WANG2023e18124,giancola2018soccernet}. While effective for their specific tasks, these specialized models lack unified reasoning capabilities and cross-modal flexibility.

\subsection{Video Understanding Using Multimodal Large Language Models}
\label{subsec:video_understanding}

Early efforts in general video understanding extended MLLMs to handle video inputs, enabling video conversation and description (e.g., Video-ChatGPT \cite{Maaz2023VideoChatGPT}, Video-LLaVA \cite{lin2023video}). However, these open source models usually fall behind SOTA commercial MLLMs, such as GPT \cite{openai2024gpt4ocard}, Gemini\cite{geminiteam2024gemini}, and Claude \cite{Claude3.5} in most video tasks, especially those that require complicated reasoning on multiple video question answer benchmarks~\cite{fang2024mmbenchvideolongformmultishotbenchmark,zhou2025mlvubenchmarkingmultitasklong}. 
Recent developments have focused on combining reinforcement learning (RL) with MLLMs, pushing models to achieve stronger reasoning abilities and demonstrate enhanced performance on various video understanding tasks, such as Video-R1\cite{feng2025video} and DeepVideo-R1 \cite{park2025deepvideor1videoreinforcementfinetuning}. Inspired by the concept of ``Thinking with images'' proposed by OpenAI~\cite{a2025_thinking}, recent works have begun using agentic reinforcement learning to train MLLM as an agent that can use tools to dynamically and iteratively interrogate visual content as part of its reasoning process~\cite{zhang2025landscapeagenticreinforcementlearning, fan2025grit, zheng2025deepeyesincentivizingthinkingimages}. This agentic concept has been extended to the temporal domain to handle long videos. 
Recent works introduce MLLMs that can ``think with long videos''~\cite{zhang2025thinkingvideosmultimodaltoolaugmented, he2025framethinkerlearningthinklong}. These models act as agents that can densely sample new video frames on demand or perform multi-turn frame spotlighting by invoking a visual toolbox. 
This allows the model to actively ``re-watch'' segments and dynamically query the video for relevant frames, drastically reducing the number of frames processed and mitigating hallucination. 

While these general-domain video MLLMs demonstrate powerful reasoning on benchmarks, such as Video-Holmes~\cite{cheng2025videoholmesmllmthinklike}, they are trained for general-purpose rather than domain-specific tasks. The current general-domain MLLMs have been shown lacking the specialized domain knowledge required to understand the intricate rules, fine-grained actions, and professional terminology of diverse sports~\cite{xia2024sportu, rao2025multi}.

\subsection{Multimodal Large Language Models in Sports}
\label{subsec:sports_models}

Existing MLLMs in sports are typically limited in one of three ways: they are (i) \textbf{training-free}, relying on static knowledge graphs with limited tasks, such as no commentary generation task, which is one of the important things for sports viewers \cite{chen2025finequestadaptiveknowledgeassistedsports}; (ii) \textbf{single-sport}, focusing on individual sports \cite{rao2025multi, you2025timesoccer, kodathala2025sv33bsportsvideounderstanding, bao2025tennistvmultimodallargelanguage}; or (iii) \textbf{single-task}, such as models designed only for commentary \cite{you2025timesoccer}.

To the best of our knowledge, a single, end-to-end trained MLLM capable of performing multiple tasks across multiple sports does not yet exist. 
Therefore, our work introduces the first MLLM to address this challenge, leveraging tool-based agentic RL during training to unify these diverse capabilities.

\begin{figure*}[htp]
    \centering
    \includegraphics[width=0.82\textwidth]{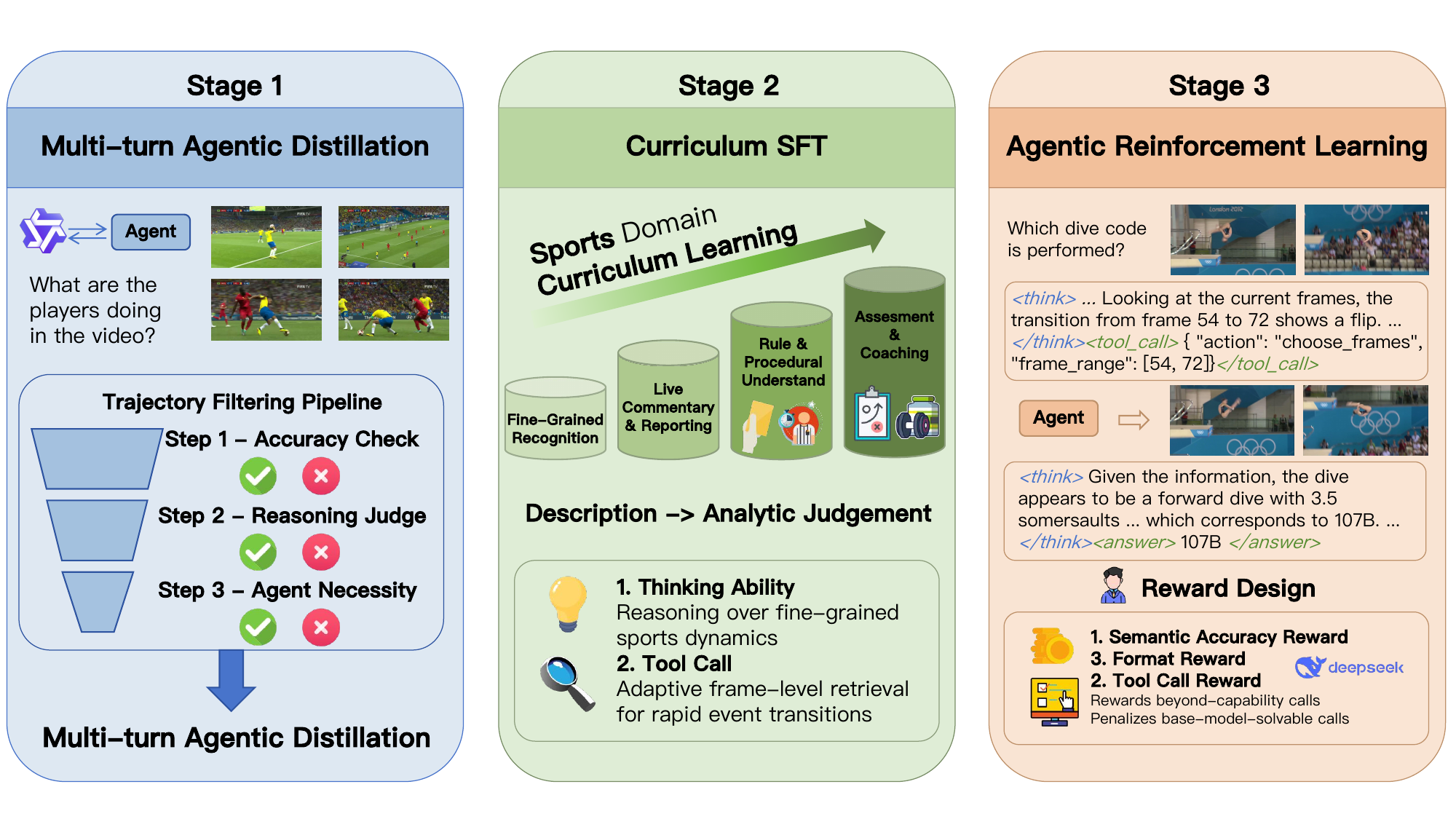}
    \caption{DeepSport training overview. Given sports videos, we first perform data distillation with a teacher MLLM to construct DeepSport-CoT data and Curriculum Supervised Fine-Tune model. We then further optimize the model with GRPO-based agentic reinforcement learning, where the agent iteratively calls a frame-extraction tool, produces chain-of-thought reasoning over new frames, and is guided by a reward manager that combines semantic accuracy, behavioral shaping, and format gating.}
    \label{pyramid}
\end{figure*}

\section{The DeepSport Framework}
\label{sec:framework}

To address the gap in multi-task, multi-sport video understanding, we introduce \textbf{DeepSport}, an end-to-end trained MLLM trained with agentic reinforcement learning with a specific tool function, that extracts related frames and performs multi-turn generation. Our framework is built upon a two-stage training pipeline: (1) a Supervised Fine-Tuning (SFT) ``cold-start'' phase, followed by (2) an Agentic Reinforcement Learning (RL) phase. This SFT+RL paradigm is a validated approach for training capable ``Thinking with Video'' agents, as seen in recent works like FrameThinker~\cite{he2025framethinkerlearningthinklong}.
Our methodology makes two primary contributions. First, we propose a \textbf{data distillation pipeline} to create a high-quality, large-scale dataset that includes the complex Chain-of-Thought (CoT) reasoning and tool-call syntax necessary for training the model. Second, we introduce a novel \textbf{tool-based reward function} within the Group Relative Policy Optimization (GRPO) framework to specifically incentivize effective multi-turn reasoning.

In this section, we first define our core paradigm (Section~\ref{subsec:paradigm}). We then detail our data distillation pipeline (Section~\ref{subsec:data_distillation}) and conclude with the two-stage training strategy (Section~\ref{subsec:training_strategy}).

\subsection{Deepsport Video Reasoning Paradigm}
\label{subsec:paradigm}

Our framework empowers the MLLM to ``think with videos'' \cite{zhang2025thinkingvideosmultimodaltoolaugmented, he2025framethinkerlearningthinklong}. The framework engages in a multi-turn reasoning loop, actively and strategically seeking information from the video for its decision-making.

The task can be defined as, given a user query $Q$ and a long video $V$, the model's reasoning process generates a trajectory $\tau$. The process is initialized at $t=1$ by providing the model with an initial context $F_1$, which is a uniformly-sampled set of $k$ frames (e.g., $k=8$) from the entire video $V$. Each frame is tagged with its \texttt{frame\_index}.

At each subsequent step $t$, the model must choose an action $A_t$ from its available action space $\mathcal{A}$, which consists of two types: 
\begin{itemize}
    \item \textbf{Frame Extraction Tool:} \texttt{<tool\_call>}
    
    \texttt{frame\_extraction\_tool(}
    \texttt{$idx_{start}, idx_{end}$)}\texttt{</tool\_call>}. 
    This is an \textit{iterative} action that allows the model to request a new set of frames from a specified temporal window.
    \item \textbf{Output Answer:} \texttt{<answer>...</answer>}. This is a \textit{terminal} action that concludes the reasoning trajectory and provides the final answer.
\end{itemize}

A complete trajectory $\tau$ is thus a sequence of $n$ steps, where each step consists of the frames provided to the model, the thought it generated, and the action it took:
\begin{equation}
\label{eq:trajectory}
\tau = \big((F_1, T_1, A_1), (F_2, T_2, A_2), \dots, (F_n, T_n, A_n)\big).
\end{equation}
Each triplet $(F_t, T_t, A_t)$ at step $t$ is defined as:
\begin{itemize}
    \item \textbf{Provided Frames ($F_t$):} The set of $k$ frames available \textit{at the beginning} of step $t$. For $t=1$, this is the initial sparse context ($F_1$). For $t>1$, this is the set of frames $F_t$ returned by the tool call $A_{t-1}$.
    
    \item \textbf{Thought ($T_t$):} The model's textual reasoning, generated within \\ \texttt{<think>...</think>} tags. The model generates this thought based on $F_t$ and all prior history.
    
    \item \textbf{Action ($A_t$):} The action chosen by the model after its thought $T_t$. If $A_t$ is the \texttt{frame\_extraction\_tool}, it generates $F_{t+1}$ for the next step. If $A_t$ is the \texttt{output\_answer}, the trajectory terminates.
\end{itemize}

Inspired by the Cognitive Consistency Verification module in FrameThinker \cite{he2025framethinkerlearningthinklong}, we apply the following method to prevent of redundant exploration. If the model executes an action $A_t$ with the \textit{exact same} temporal interval as its preceding action $A_{t-1}$, the framework treats this as an invalid ``format error,'' which is penalized during training.

\subsection{Data Distillation Pipeline}
\label{subsec:data_distillation}

A significant bottleneck in training a sport MLLMs for iterative reasoning is the lack of high-quality, large-scale sports-specific SFT data that demonstrates the desired step-by-step reasoning and tool-use behavior.

\noindent\textbf{1. Data Generation.}
We first curated a diverse set of 9 existing works (Table~\ref{tab:datasets}) with 10 different data sources to cover a wide spectrum of \textbf{12 distinct sports}, spanning team sports (Soccer, Basketball, Volleyball, American Football, Ice Hockey, Baseball), racket games (Table Tennis, Badminton), combat sports (Fencing, Boxing), and artistic disciplines (Diving, Gymnastics). We unified these disparate data to establish a comprehensive taxonomy of tasks across four high-level dimensions: 
(1) \textbf{Fine-Grained Recognition}: encompassing foundational visual perception tasks such as Action Sequencing, Player Identification, Camera Transition Analysis, etc.; 
(2) \textbf{Rule \& Procedural Logic}: focusing on regulatory knowledge, including Foul Classification, Referee Decision Analysis, Penalty Description, etc.; 
(3) \textbf{Assessment \& Coaching}: requiring expert-level reasoning for Tactical Analysis, Technical Error Identification, Scoring Prediction, etc.; (4) \textbf{Live Commentary \& Reporting}: covering multimodal narration tasks like Play-by-Play Commentary, Game Situation Analysis, Score Reading, etc.

Since several of these benchmarks are not in a question-answering format (FineDiving~\cite{xu2022finediving} for action quality, SoccerReplay-1988 \cite{rao2024towards} for commentary generation, FACTS \cite{lai2024facts} for action classification, or T3Set \cite{10.1145/3711896.3737407} for tactic correction), our first step was to unify them. We designed a series of task-specific templates to convert their original labels into structured QA pairs. 

\begin{table}[h]
\centering
\caption{Source datasets curated for our data distillation pipeline. Tasks marked with (*) were converted into a QA format using templates.}
\label{tab:datasets}
\resizebox{0.65\columnwidth}{!}{
\renewcommand{\arraystretch}{1.5} 
\begin{tabular}{@{}lll@{}}
\toprule
\textbf{Dataset Source} & \textbf{Sport(s)} & \textbf{Original Task / Content} \\
\midrule
SoccerBench \cite{rao2025multi} & Soccer &  Multi-task QA \\
SoccerReplay-1988 (\cite{rao2024towards}) & Soccer & Commentary Generation (*) \\
SPORTU \cite{xia2024sportu} & Multi-sport & Multi-level \& -task QA  \\
Sports-QA \cite{li2024sports} & Multi-sport &  Multi-task QA \\
SportR \cite{xia2025sportrbenchmarkmultimodallarge} & Multi-sport & Foul \& Tactic Reasoning QA \\
FineDiving \cite{xu2022finediving} & Diving & Action Quality Assessment (*) \\
FACTS \cite{lai2024facts} & Fencing, Boxing & Action Classification (*) \\
T3-Set (\cite{10.1145/3711896.3737407}) & Table Tennis & Tactic Correction (*) \\
X-VARS(\cite{held2024x}) & Soccer & Foul Recognition \& Explanation\\
\bottomrule
\end{tabular}
}
\end{table}

\noindent\textbf{2. Data Splitting and Integrity.}
We follow video-level splitting: all questions associated with a single video (even if there are multiple) are assigned exclusively to one data split. This prevents the model from being trained and tested on frames from the same video. Furthermore, we explicitly handle dataset overlap; for instance, any video clip used in the SoccerBench~\cite{rao2025multi} test set is excluded from our SoccerReplay-1988~\cite{rao2024towards} training pool. The detailed number of QA split will be illustrated in Section~\ref{subsec:exp_setup}. 

\noindent\textbf{3. CoT Data Distillation.}
With a unified set of QA prompts, we generate high-quality reasoning trajectories. We prompt the current SOTA model, Qwen3-VL-235B-A22B-Thinking~\cite{qwen3technicalreport}, to generate detailed, step-by-step reasoning. Critically, the prompt instructs the model to invoke the \texttt{frame\_extraction\_tool} (defined in Section~\ref{subsec:paradigm}) when necessary. This process results in CoT trajectories that are interleaved with tool calls.

To ensure the data quality and prevent reasoning hallucinations or redundant tool usage, we propose a rigorous three-step filtering pipeline: \textbf{Step 1 - Accuracy Check:} We use DeepSeek-V3.2-Exp~\cite{deepseekai2025deepseekv32pushingfrontieropen} to verify if the model's predicted final answer matches the ground truth. \textbf{Step 2 - Logic Consistency Review:} We employ DeepSeek-V3.2~\cite{deepseekai2025deepseekv32pushingfrontieropen} to evaluate the reasoning trace. This ensures the reasoning is self-consistent and that any revision of previous conclusions is justified by new information, strictly penalizing unprompted contradictions. \textbf{Step 3 - Retrieval Usefulness Evaluation:} We utilize a Kimi K2.5~\cite{kimiteam2026kimik25visualagentic} to assess the visual necessity of the tool call. It verifies whether the dynamically retrieved frames provide genuinely new, finer-grained visual details compared to the initial frames, ensuring the retrieval is not redundant.

Only trajectories that successfully pass all three verification steps are retained. As a result, we constructed the \textbf{DeepSport-CoT-14K} dataset, which contains 14,599 high-quality Q\&A pairs with robust CoT annotations, to be used later in SFT training.

\subsection{Two-Stage Training Strategy}
\label{subsec:training_strategy}

 We first perform Supervised Fine-Tuning (SFT) to teach the model the CoT format as the ``cold start'', which has been wildly adopted in recent works~\cite{deepseekai2025deepseekr1incentivizingreasoningcapability, feng2025video,zhang2025thinkingvideosmultimodaltoolaugmented}, followed by a Reinforcement Learning (RL) phase to optimize the model's reasoning and tool-use policy.

\subsubsection{Sports Curriculum SFT Cold-Start}
\label{subsec:sft}
We first perform a Supervised Fine-Tuning (SFT) phase using the \textbf{DeepSport-CoT-14K} dataset. The objective of this phase is twofold: (1) to serve as a ``cold-start'' to familiarize the model with the complex CoT trajectory structure (i.e., learning the syntax of \texttt{<think>...} and \texttt{<tool\_call>...}), and (2) to incrementally build the model's domain knowledge. 

To achieve this, we introduce a \textbf{Sports Curriculum Learning} strategy. The underlying rationale is that high-level analytical tasks (e.g., Live Commentary, Assessment \& Coaching) are fundamentally predicated on solid foundational perception (Fine-Grained Recognition). Therefore, we partition the SFT samples into 5 progressive stages. As training advances from Stage 1 to Stage 5, we gradually decrease the proportion of Fine-Grained Recognition tasks (from 79\% down to 47\%) while steadily increasing the ratio of complex reasoning tasks, such as Rule \& Procedural Logic (from 7\% to 20\%) and Assessment \& Coaching (from 5\% to 21\%). This progressive data curriculum ensures that the model masters basic visual grounding before tackling expert-level logic. This initial alignment is crucial for stabilizing the subsequent RL phase, providing a reasonable policy capable of producing valid actions.

\subsubsection{Agentic RL with Gated Tool Rewards}
\label{subsec:agentic_rl}

In the second stage, we optimize the model's policy using Reinforcement Learning.

\paragraph{Group Relative Policy Optimization (GRPO).}
During the reinforcement learning phase, we employ the Group Relative Policy Optimization (GRPO)~\cite{shao2024deepseekmath}, which has been commonly used in reinforcement learning for model training across multiple domains~\cite{deepseekai2025deepseekr1incentivizingreasoningcapability,liu2025finr1largelanguagemodel,feng2025video}. For each prompt $q$ from the RL set, we sample a group of $G$ trajectories $\{\tau_i\}_{i=1}^G \sim \pi_{\text{old}}(\cdot\mid q)$ and obtain rewards $\{r_i\}_{i=1}^G$. The advantage $A_i$ is defined by the group statistics:
\begin{equation}
\label{eq:advantage}
A_i \;=\; \frac{r_i - \mu_r}{\sigma_r}\!,
\end{equation}
where $\mu_r$ and $\sigma_r$ are the mean and standard deviation of the rewards $\{r_i\}$. Letting $\rho_i = \frac{\pi_\theta(\tau_i\mid q)}{\pi_{\text{old}}(\tau_i\mid q)}$, the GRPO objective is:
\begin{equation}
\label{eq:grpo_full}
\begin{split}
\mathcal{J}_{\text{GRPO}}(\theta) = \mathbb{E}\Bigg[ \frac{1}{G} \sum_{i=1}^{G} &\min\Big(\rho_i A_i, \operatorname{clip}(\rho_i, 1-\varepsilon, 1+\varepsilon) A_i\Big) \\
&- \beta \cdot \text{KL}\big(\pi_\theta(\cdot \mid q) \| \pi_{\text{ref}}(\cdot \mid q)\big) \Bigg].
\end{split}
\end{equation}
where $\pi_{\text{ref}}$ is the reference model, which is usually the initial SFT model.

\paragraph{Gated Tool-Use Reward Function.}
\label{par:reward}
To train our model, we design a multi-component reward function $R(\tau)$ that encourages task success, selective tool invocation, and well-formed outputs.

\noindent\textbf{1. Sample Classification.}
A naïve tool-use bonus would incentivize the model to invoke the \texttt{frame\_extraction\_tool} indiscriminately, even when the initial frames already suffice. To prevent this reward hacking, we pre-classify every training sample by evaluating the base model on the initial uniformly-sampled frames \textit{before} RL training:
\begin{itemize}
    \item \textbf{Class-A}: samples the base model \textit{fails} to answer correctly from the initial frames, indicating that temporal re-inspection is likely necessary.
    \item \textbf{Class-B}: samples the base model \textit{can} answer correctly from the initial frames, indicating that further tool invocation is unnecessary.
\end{itemize}
We denote the class of a trajectory $\tau$ as $c(\tau)\in\{\texttt{a},\texttt{b}\}$.

\noindent\textbf{2. Core Objective.} The primary reward signal $R_{\text{acc}}$ measures task success via an LLM-as-a-judge semantic similarity score:
\begin{equation}
R_{\text{acc}}(\tau) = \mathrm{acc}(\tau), \quad \mathrm{acc}(\tau) \in [0,1].
\label{eq:acc}
\end{equation}

\noindent\textbf{3. Selective Tool-Use Incentive.}
For \textbf{Class-A} samples (initially unanswerable), the model is rewarded for using the tool when it leads to a correct answer; for \textbf{Class-B} samples (initially answerable), the model is rewarded for \textit{refraining} from tool use while still answering correctly:
\begin{equation}
R_{\text{tool}}(\tau) =
\begin{cases}
\dfrac{\mathrm{acc}(\tau)}{2}, & \text{if } c(\tau){=}\texttt{a},\; \text{used tool},\; \mathrm{acc}(\tau){\ge}0.5, \\
\dfrac{\mathrm{acc}(\tau)}{2}, & \text{if } c(\tau){=}\texttt{b},\; \text{no tool},\; \mathrm{acc}(\tau){\ge}0.5, \\
0, & \text{otherwise.}
\end{cases}
\label{eq:tool}
\end{equation}
This asymmetric design closes the reward hacking loophole: the model cannot earn $R_{\text{tool}}$ by blindly invoking the tool on every sample, as doing so on Class-B samples yields zero bonus. Instead, the model must learn to \textit{judge whether the initial frames are sufficient} before deciding to act.

\noindent\textbf{4. Format Gating and Total Reward.}
We enforce the structural integrity of the trajectory via a format validation that checks matched and correctly nested \texttt{<think>}, \texttt{<tool\_call>}, and \texttt{<answer>} tags, non-empty tag contents, and valid tool-call arguments (e.g., strictly narrowing frame ranges with no redundant re-requests). The total reward is:
\begin{equation}
R(\tau) =
\begin{cases}
R_{\text{acc}}(\tau) + R_{\text{tool}}(\tau), & \text{if format is valid,} \\[4pt]
-0.05, & \text{otherwise.}
\end{cases}
\label{eq:total}
\end{equation}

\section{Experiments}
\label{sec:experiments}
\subsection{Experimental Setup}
\label{subsec:exp_setup}

\paragraph{Datasets.}
As described in Section~\ref{subsec:data_distillation}, our data pipeline yields three distinct subsets based on a strict video-level split. First, we construct \textbf{DeepSport-CoT-14k}, consisting of 14k high-quality, LLM-filtered CoT QA pairs used specifically for the SFT cold-start phase. Second, for the reinforcement learning phase, we utilize \textbf{DeepSport-RL-63k}, which comprises 63k original, unified QA pairs (prompt plus ground truth answer) serving as the environment prompts. Finally, we evaluate performance on our held-out Testing Benchmark, containing 6.7k QA pairs that cover all tasks and sports to ensure a rigorous evaluation.

\paragraph{Implementation Details.}
We select Qwen2.5-VL-7B~\cite{bai2025qwen2} as our backbone model, processing video inputs at a resolution of $640 \times 360$. Our training pipeline consists of two stages: we first fine-tune the model on the DeepSport-CoT-14k set for 1 epoch \textbf{following the 5-stage curriculum learning design}, followed by reinforcement learning via our tool-based GRPO algorithm with a batch size of 32 and a rollout number of 8 on the DeepSport-RL-63k set for 300 steps. The model is trained to generate up to 12,800 tokens per response. Training was performed on 8 $\times$ H20 GPUs for totaling 796 GPU hours.

\paragraph{Baselines.}
We compare DeepSport against a comprehensive suite of MLLMs using uniform 16-frame sampling per video, including both open- and closed-source models . Specifically, we select models from the Qwen family (Qwen2.5-VL-7B-Instruct as our primary baseline for direct comparison, Qwen3-VL-8B-Instruct/Thinking, and Qwen3-VL-235B-A22B-Thinking~\cite{bai2025qwen2, qwen3technicalreport}), as well as InternVL3.5-14B~\cite{wang2025internvl3_5} and Video-R1~\cite{feng2025video}. In addition, we evaluate DeepSport under alternative frame-sampling strategies, including T$^*$~\cite{ye2025trethinkingtemporalsearch} and VideoITG~\cite{wang2025videoitgmultimodalvideounderstanding}, and we further compare against the soccer-specific model SoccerChat~\cite{gautam2025soccerchatintegratingmultimodaldata}.

\subsection{Main Results}

We evaluate models on our 6.7k multi-task, multi-sport test set using an LLM-as-Judge (Claude 4.5 Sonnet~\cite{anthropic_2025_introducing}), with scores scaled to a percentage format.

The comprehensive evaluation results are presented in Table~\ref{tab:model_sports_performance}. This trend of high performance extends across the diverse task spectrum of our benchmark. When viewed holistically, DeepSport's well-rounded capabilities become clear. It achieves an overall average score of 37.67, establishing a new state-of-the-art by surpassing the powerful closed-source GPT-5\cite{singh2025openaigpt5card} and the massive Qwen3-VL-235B-A22B-Thinking.

\begin{table*}[t]
\centering
\caption{Performance comparison on the test set. We evaluate models across four high-level dimensions designed to assess holistic sports tasks. Each dimension encompasses diverse specific capabilities: Fine-Grained Recognition involves visual perception tasks (e.g., Action Recognition, Player Identification); Rule \& Procedural tests regulatory knowledge (e.g., Foul Classification, Referee Decision Analysis); Assessment \& Coaching evaluates expert reasoning (e.g., Tactical Analysis, Scoring Prediction); and Live Commentary \& Reporting assesses multimodal narration (e.g., Play-by-Play Commentary, Score Reading). \colorbox{red!15}{\textbf{Bold}} indicates the best performance, and \colorbox{blue!15}{blue background} indicates the second best. Green arrows (\textcolor{green!60!black}{$\uparrow$}) denote the absolute performance gains of our DeepSport over the Qwen2.5-VL-7B baseline.}
\label{tab:model_sports_performance}
\resizebox{0.85\textwidth}{!}{%
\large
\renewcommand{\arraystretch}{1.3}
\begin{tabular}{l|c|cccc|c}
\toprule
\textbf{Model} & \textbf{Frames} & \textbf{Fine-Grained} & \textbf{Rule \&} & \textbf{Assessment \&} & \textbf{Live Commentary} & \textbf{Overall} \\
& & \textbf{Recognition} & \textbf{Procedural} & \textbf{Coaching} & \textbf{\& Reporting} & \textbf{Avg.} \\
\midrule
GPT-5 & 16 & 46.50 & \cellcolor{blue!15}32.89 & \cellcolor{blue!15}27.01 & 22.76 & \cellcolor{blue!15}35.70 \\
Qwen3-VL-8B-Instruct & 16 & 43.82 & 23.49 & 25.52 & 20.77 & 31.66 \\
Qwen3-VL-8B-Thinking & 16 & \cellcolor{blue!15}46.75 & 31.98 & 23.74 & 21.96 & 34.83 \\
Video-R1 & 16 & 38.21 & 17.76 & 13.73 & 10.16 & 24.06 \\
InternVL3.5-14B & 16 & 46.38 & 29.10 & 22.64 & 17.59 & 33.01 \\
Qwen3-VL-235B-A22B-Thinking & 16 & 44.96 & 31.03 & \cellcolor{red!15}\textbf{28.72} & \cellcolor{red!15}\textbf{25.59} & 35.36 \\
T$^*$ & 16 & 35.79 & 19.49 & 9.57 & 7.56 & 22.30 \\
VideoITG & 16 & 35.39 & 15.25 & 9.64 & 5.96 & 20.85 \\
SoccerChat & 16 & 31.43 & 29.20 & 8.69 & 10.23 & 23.19 \\
\midrule
Qwen2.5-VL-7B-Instruct (Baseline) & 16 & 28.20 & 13.71 & 9.09 & 3.13 & 16.98 \\
Baseline + SFT (Ours)& 10 & 40.67 & 24.38 & 21.40 & 21.07 & 29.91 \\
Baseline + Sports Curriculum SFT (Ours) & 8.72 & 46.11 & 32.61 & 17.87 & 17.70 & 32.48 \\
\textbf{DeepSport (Ours)} & 9.81 
  & \cellcolor{red!15}\textbf{49.89}{\textcolor{green!60!black}{\small{$\uparrow$\!21.69}}} 
  & \cellcolor{red!15}\textbf{41.20}{\textcolor{green!60!black}{\small{$\uparrow$\!27.49}}} 
  & 19.19{\textcolor{green!60!black}{\small{$\uparrow$\!10.10}}} 
  & \cellcolor{blue!15}22.93{\textcolor{green!60!black}{\small{$\uparrow$\!19.80}}} 
  & \cellcolor{red!15}\textbf{37.67}{\textcolor{green!60!black}{\small{$\uparrow$\!20.69}}} \\
\bottomrule
\end{tabular}%
}
\end{table*}

Quantitatively, DeepSport's superiority is most pronounced in perception-heavy and rule-bound tasks. It secures a dominant lead in Fine-Grained Recognition with a score of 49.89 and Rule \& Procedural tasks with 41.20. This substantial margin over generalist models validates our core hypothesis: the active tool-use paradigm effectively captures fleeting visual details missed by passive processors, while agentic RL successfully internalizes complex domain-specific regulations.

To validate our progressive training paradigm, we conducted an ablation study on the training stages (shown in the lower section of Table~\ref{tab:model_sports_performance}). While direct SFT yields a significant performance bump, applying our staged progressive training (Sports Curriculum SFT) further elevates the overall average score from 29.91 to 32.48. Notably, this curriculum drives massive gains in foundational perception (Fine-Grained Recognition improves from 40.67 to 46.11) and regulatory logic (Rule \& Procedural improves from 24.38 to 32.61). This clearly demonstrates that mastering basic visual grounding before tackling expert-level logic is a highly effective strategy for sports video understanding.

In terms of efficiency, while the Qwen3-VL-235B-A22B-Thinking model retains a marginal edge in generative tasks like \textit{Assessment \& Coaching} and \textit{Live Commentary}—likely leveraging its massive pre-trained knowledge base—DeepSport delivers highly competitive performance despite having only $\sim$3\% of the parameter count. This signifies that our distillation pipeline effectively transfers high-level reasoning capabilities to a compact backbone. Moreover, our model achieves the SOTA overall score using only 9.81 frames on average, significantly fewer than the fixed 16 frames used by other models.

\begin{figure*}[htp]
    \centering
    \includegraphics[width=0.8\textwidth]{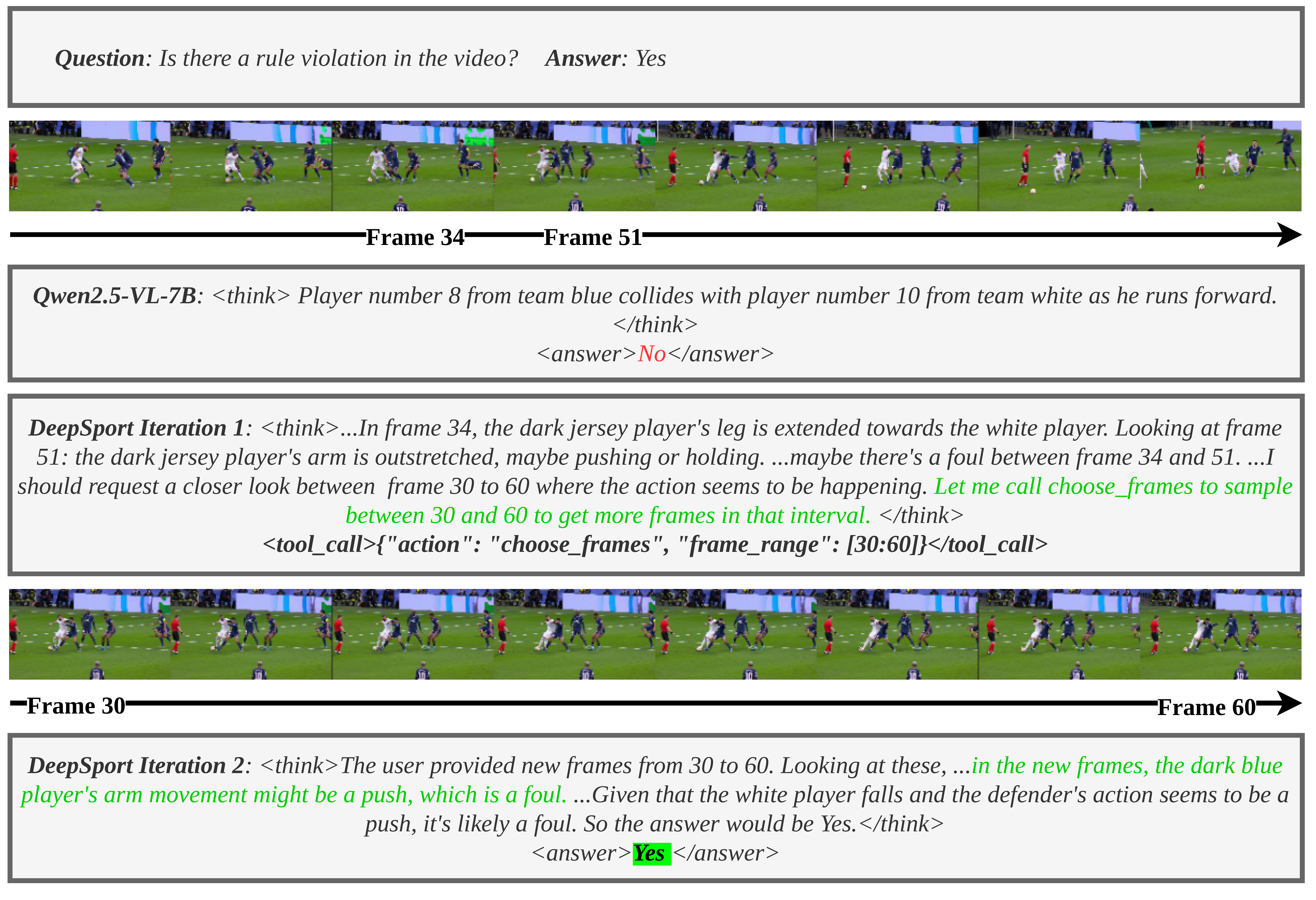}
    \caption{We present a comparison on the foul detection task, where the model need capture fine-grained movement. The Qwen2.5-VL-7B-Instruct model, relying on passive, single-pass processing of 16 sparsely sampled frames, misses the high-speed contact and made the wrong conclusion. Our DeepSport model activates the multi-turn conversation by invoking the frame\_extraction\_tool(30, 60) to retrieve second-round relevant frames. Our model correctly identifies the foul.}
\label{fig:american}
\end{figure*}

Furthermore, the necessity of specialized training is underscored by the comparison with general-domain baselines. DeepSport achieves a score improvement of 20.69 over its own backbone, Qwen2.5-VL-7B, and significantly outperforms the general RL-trained Video-R1 model. This performance gap underscores that general-domain training fails to capture sport-specific nuances, highlighting the critical need for our multi-task specialized model. Figure~\ref{fig:american} shows the output comparison between our DeepSport model and the backbone model, Qwen2.5-VL-7B.

\subsection{Generalization on General Benchmarks}
\label{subsec:generalization}

To investigate whether our sport-specific training compromises broader video understanding capabilities, we expanded our evaluation beyond sports to encompass both general long video benchmarks and fine-grained action recognition datasets. Specifically, we evaluate DeepSport on two distinct dimensions: (1) long-form video understanding using LVBench~\cite{zhou2024mlvu}, LongVideoBench~\cite{wu2024longvideobenchbenchmarklongcontextinterleaved}, and Video-MME (Long)~\cite{fu2025videommefirstevercomprehensiveevaluation}; and (2) fine-grained action and motion understanding using DREAM-1K~\cite{wang2024tarsierrecipestrainingevaluating}, MotionBench~\cite{hong2025motionbenchbenchmarkingimprovingfinegrained}, and ActionAtlas~\cite{salehi2024actionatlasvideoqabenchmarkdomainspecialized}.

As shown in Table~\ref{tab:general_bench}, DeepSport maintains competitive performance on general video tasks. While there is a slight performance drop on VideoMME (Long) compared to its backbone, Qwen2.5-VL-7B-Instruct, a common trade-off in domain-specific fine-tuning~\cite{luo2025empirical}, DeepSport improves upon the baseline in both LVBench and LongVideoBench. Most notably, it achieves these robust results while processing significantly fewer frames ($\sim$13 frames) compared to the fixed 32 frames of the backbone, demonstrating the cross-domain efficiency of our active frame extraction mechanism.

Furthermore, DeepSport exhibits substantial improvements in general action recognition tasks. On broader motion benchmarks like DREAM-1K and MotionBench, DeepSport significantly outperforms the baseline by effectively capturing fleeting motion dynamics through its iterative, multi-turn frame sampling. To further test the generalization limits within the sports domain, we specifically evaluated DeepSport on unseen sports in the ActionAtlas benchmark. We purposefully selected sports that were entirely excluded from our training set. DeepSport still surpasses the baseline model. This strongly indicates that our model does not merely overfit to or memorize sport-specific rules; instead, it successfully internalizes the underlying interconnectedness of athletic human movements, enabling strong transferability to novel disciplines.

\begin{table*}[t]
\centering
\caption{Performance on general video understanding and action recognition benchmarks. DeepSport maintains competitive performance on long videos while significantly improving general and unseen sports action recognition, all with fewer average frames.}
\label{tab:general_bench}
\resizebox{0.85\textwidth}{!}{%
\renewcommand{\arraystretch}{1.2}
\begin{tabular}{l|cc|cc|cc|cc|cc|cc}
\toprule
\multirow{3}{*}{\textbf{Model}} & \multicolumn{6}{c|}{\textbf{General Video Understanding}} & \multicolumn{6}{c}{\textbf{Action \& Motion Recognition}} \\
\cline{2-13}
 & \multicolumn{2}{c|}{\textbf{LVBench}} & \multicolumn{2}{c|}{\textbf{LongVideoBench}} & \multicolumn{2}{c|}{\textbf{VideoMME (Long)}} & \multicolumn{2}{c|}{\textbf{DREAM-1K}} & \multicolumn{2}{c|}{\textbf{MotionBench}} & \multicolumn{2}{c}{\textbf{ActionAtlas}} \\
 & \textbf{Acc} & \textbf{Frames} & \textbf{Acc} & \textbf{Frames} & \textbf{Acc} & \textbf{Frames} & \textbf{F1} & \textbf{Frames} & \textbf{Acc} & \textbf{Frames} & \textbf{Acc} & \textbf{Frames} \\
\midrule
Qwen2.5-VL-7B-Instruct & 31.6 & 32 & 43.2 & 32 & \textbf{41.6} & 32 & 26.0 & 16.0 & 43.1 & 16.0 & 25.6 & 16.0 \\
\textbf{DeepSport (Ours)} & \textbf{32.0} & \textbf{13.2} & \textbf{45.9} & \textbf{12.7} & 40.4 & \textbf{12.5} & \textbf{30.5} & \textbf{8.0} & \textbf{48.5} & \textbf{8.3} & \textbf{27.2} & \textbf{8.5} \\
\bottomrule
\end{tabular}%
}
\end{table*}

\subsection{Error Analysis}
\label{subsec:error_analysis}

To investigate the limitations of DeepSport, we conducted a manual inspection of 70 randomly sampled failure cases. 
We classify errors based on the the first and primary error the model made, categorizing them into six primary types:
(1) \textbf{Tool Grounding Failure (42.9\%):} The model invoked the tool but retrieved a temporal window that missed the critical event, making the answer impossible.
(2) \textbf{Visual Hallucination (37.1\%):} The model successfully retrieved the correct frames but misidentified objects, actions, or attributes.
(3) \textbf{Domain Knowledge Gap (11.4\%):} The model perceived the visual information correctly but lacked the specific sports terminology to describe it.
(4) \textbf{Rule Misapplication (4.3\%):} The model understood the visual scene and terms but failed in the logical reasoning required to apply the rule.
(5) \textbf{Internal Consistency Error (2.9\%):} The model's reasoning contradicted its final answer or violated the output format.
(6) \textbf{Incomplete Exploration (1.4\%):} The model failed to invoke the frame extraction tool despite the query requiring additional visual evidence.

\begin{figure}[t]
    \centering    \includegraphics[width=0.55\linewidth]{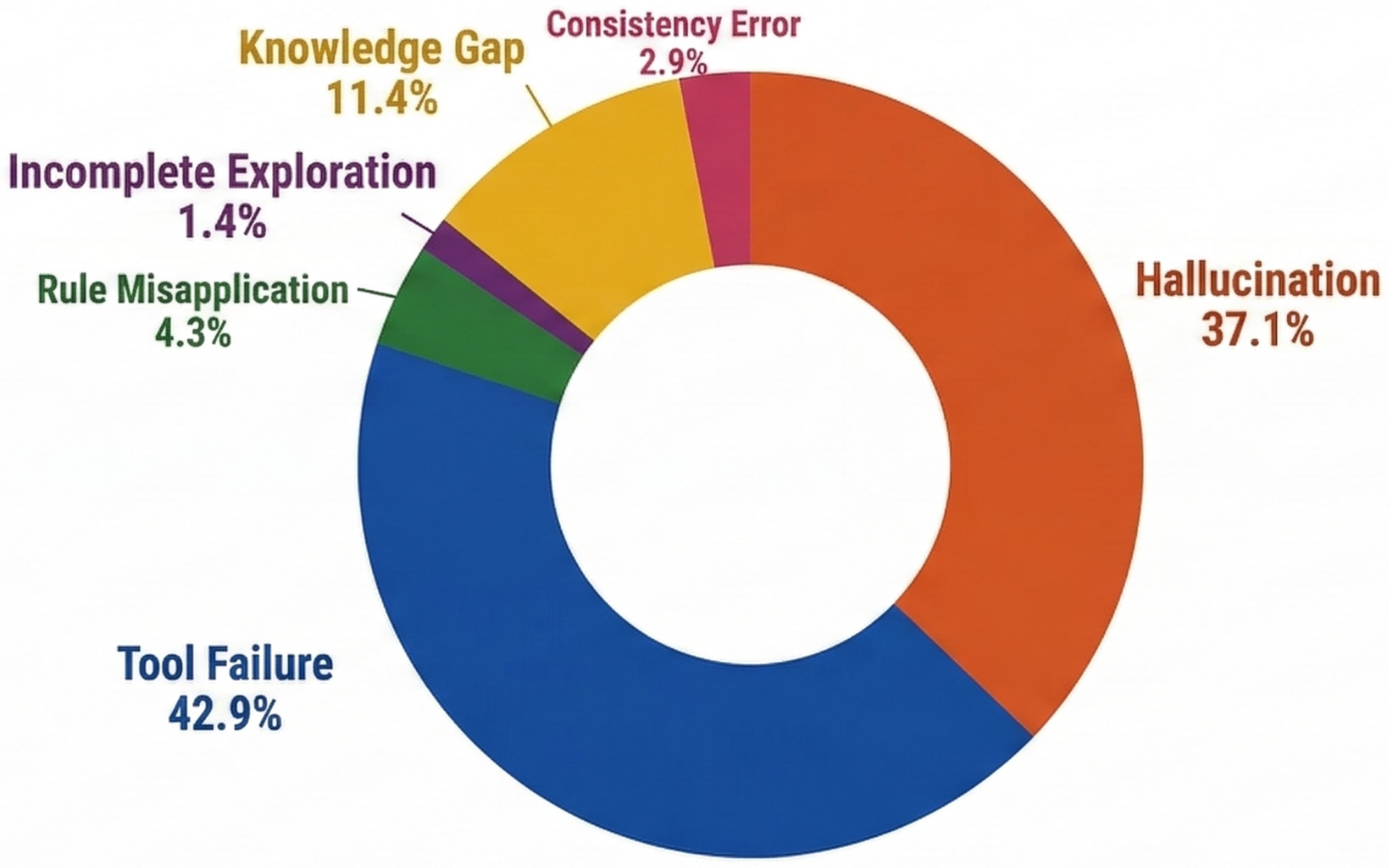} 
    \caption{Error analysis. Tool Grounding Failure highlights that precise temporal localization remains the primary bottleneck, followed by fine-grained Visual Hallucination.}
    \label{fig:error_analysis}
\end{figure}

As shown in Figure~\ref{fig:error_analysis}, the dominance of Tool Failure indicates that while our model is capable of reasoning, the primary bottleneck remains accurate \textit{temporal localization}. Furthermore, the high rate of Visual Hallucination suggests that future work should focus on enhancing the fine-grained visual ability.
\section{Conclusion}
\label{sec:conclusion}

In this paper, we introduced \textbf{DeepSport}, the first end-to-end trained MLLM model designed to unify multi-task video understanding across diverse sports. By synergizing a data distillation pipeline with tool-based agentic reinforcement learning, DeepSport shifts the paradigm from passive frame processing to active, iterative multi-turn reasoning. Our extensive experiments demonstrate that this approach significantly outperforms existing state-of-the-art models, establishing a new benchmark for sport-specific video understanding.

\paragraph{Limitations and Future Work.}
Despite these advancements, our work also highlights the inherent limitations of the current sports AI landscape, primarily stemming from data sparsity. While we have curated a comprehensive set of tasks to date, the availability of high-quality, publicly accessible data remains uneven across different sports. For instance, while soccer benefits from abundant commentary and event data, niche sports like fencing or diving often lack such dense, linguistic annotations, making tasks like ``fencing commentary generation'' currently infeasible. While our framework currently covers a representative range of major sports, numerous less-popular ones remain uncovered.

Second, our error analysis reveals that Tool Grounding Failure remains the primary issue, indicating that while the model effectively decides when to use tools, its ability to pinpoint exact temporal windows needs further refinement. Future work will focus on improving the temporal precision of the retrieval module and developing more comprehensive, multi-granular datasets to pave the way for a truly universal sports intelligence.

%
%
\bibliographystyle{splncs04}
\bibliography{main}
\end{document}